\documentclass[12pt,halfline,a4paper]{ouparticle}

\usepackage{graphicx}      
\usepackage{amsmath}       
\usepackage{amsthm}
\usepackage{amsfonts}
\usepackage{bm}            
\usepackage[round]{natbib} 
\usepackage{color}         
\usepackage{ulem}          

\usepackage{caption}
\usepackage{subcaption}

\begin{document}

\title{Machine Learning for Stochastic Parametrisation}

\author{%
\name{Hannah M. Christensen, Salah Kouhen, Greta Miller,}
\address{Atmospheric, Oceanic
and Planetary Physics, Dept. of Physics, \\
University of Oxford, Oxford OX1 3PU, UK}
\and
\name{and Raghul Parthipan}
\address{Dept. of Computer Science, University of Cambridge, \\ Cambridge, CB3 0FD, UK, \\ and British Antarctic Survey, Cambridge, CB3 0ET, UK}
\and
\email{Hannah.Christensen@physics.ox.ac.uk}
}

\abstract{Atmospheric models used for weather and climate prediction are traditionally formulated in a deterministic manner. In other words, given a particular state of the resolved scale variables, the most likely forcing from the sub-grid scale processes is estimated and used to predict the evolution of the large-scale flow. However, the lack of scale-separation in the atmosphere means that this approach is a large source of error in forecasts. Over recent years, an alternative paradigm has developed: the use of stochastic techniques to characterise uncertainty in small-scale processes. These techniques are now widely used across weather, sub-seasonal, seasonal, and climate timescales. In parallel, recent years have also seen significant progress in replacing parametrisation schemes using machine learning (ML). This has the potential to both speed up and improve our numerical models. However, the focus to date has largely been on deterministic approaches. In this position paper, we bring together these two key developments, and discuss the potential for data-driven approaches for stochastic parametrisation. We highlight early studies in this area, and draw attention to the novel challenges that remain.}

\date{\today}


\maketitle

\newpage

\section*{Impact Statement}
Weather and climate predictions are relied on by users from industry, charities, governments, and the general public. The largest source of uncertainty in these predictions arises from approximations made when building the computer model used to make them. In particular, the representation of small-scale processes such as clouds and thunderstorms is a large source of uncertainty because of their complexity and their unpredictability. Machine learning (ML) approaches, trained to mimic high-quality datasets, present an unparalleled opportunity to improve the representation of these small-scale processes in models. However, it is important to account for the unpredictability of these processes while doing so. In this paper, we demonstrate the untapped potential of such probabilistic ML approaches for improving weather and climate prediction.

\section{Introduction} \label{sec:intro}

Weather and climate models exhibit long-standing biases in mean state, modes of variability, and the representation of extremes. These biases hinder progress across the World Climate Research Programme grand challenges. Understanding and reducing these biases is a key focus for the research community.

At the heart of weather and climate models are the physical equations of motion which describe the atmosphere and ocean systems. To predict the evolution of the climate system, these equations are discretised in space and time. The resolution ranges from order 100 km and 30-60 minutes in a typical climate model, through 10 km and 5-10 minutes in global numerical weather prediction (NWP) models, to one km and a few tens of seconds for state-of-the-art convection permitting runs. The impact of unresolved scales of motion on the resolved scales is represented in models through \emph{parametrisation schemes} \citep{christensen2022}. Many of the biases in weather and climate models stem from the assumptions and approximations made during this parametrisation process \citep{Hyder2018}. Furthermore, despite their approximate nature, conventional parametrisation schemes account for twice as much compute time as the dynamical core in a typical atmospheric model \citep{Wedi2013}.

Replacing existing parametrisation schemes with statistical models trained using machine learning (ML) has great potential to improve weather and climate models, both in terms of reduced computational cost and increased accuracy. For example, ML has been used to emulate existing parametrisation schemes including radiation \citep{chevallier1998,krasnopolsky2005,ukkonen2020} and convection \citep{ogorman2018using}, realising speed-ups of up to 80 times in the case of radiation. By emulating more complex (and therefore expensive) versions of a parametrisation scheme, the accuracy of the weather or climate model can also be improved compared to the control simulation, at minimal computational cost \citep{Chantry2021,gettelman2021}. The accuracy of the climate model can be further improved by training ML models on high fidelity data sets. For example, several authors have coarse grained high resolution models which explicitly resolve deep convection to provide training data for ML models, leading to improved parametrisations \citep{gentine2018could,brenowitz2020}. Recent research has focused on challenges including online stability \citep{brenowitz2018prognostic,brenowitz2020,yuval2020,yuval2021}, conservation properties \citep{Beucler2021}, and the ability of ML emulators to generalise and perform well in climate change scenarios \citep{Beucler2021}.


The schemes referenced above all implicitly assume that the grid-scale variables of state fully determine the parametrised impact of the sub-grid scales \citep{Palmer_2019}. This assumption is flawed. An alternative approach is \emph{stochastic parametrisation} \citep{leutbecher2017}, where the sub-grid tendency is drawn from a probability distribution conditioned on the resolved scale state. Stochastic parametrisation schemes are widely used in the weather and subseasonal-to-seasonal forecasting communities, where they have been shown to improve the reliability of forecasts \citep{leutbecher2017}. Furthermore, the inclusion of stochastic parametrisations from the weather forecasting community into climate models has led to dramatic improvements in longstanding biases \citep{berner2017,Christensen2017a}. While rigorous theoretical ideas exist to demonstrate the need for stochasticity in fluid dynamical models \citep{Gottwald_2017}, the approaches currently used tend to be pragmatic and ad hoc in their formulation \citep{berner2017}. There is therefore great potential for data-driven approaches in this area, in which the uncertainty characteristics of sub-grid scale processes are derived from observational or high-resolution model data sets \citep{christensen2020}. In this article, we discuss the potential for machine learning to transform stochastic parametrisation. In section \ref{sec:stoch} we provide the physical motivation for stochastic parametrisations, and give an overview of their benefits and limitations. In section \ref{sec:ml} we discuss the potential for machine learning and highlight some preliminary studies in this space. We conclude in section \ref{sec:conc} by issuing an invitation to the ML parametrisation community to move towards such a probabilistic framework.

\section{Stochastic Parametrisation} \label{sec:stoch}

\subsection{Deterministic versus stochastic closure} \label{sec:closure}


The conceptual framework utilised by the ML parametrisations typically mirrors that of the schemes they replace. 
The Navier-Stokes and thermodynamic equations which describe the atmosphere (and ocean) are discretised onto a spatial grid. The resultant grid-scale variables of state (e.g. temperature, horizontal winds, and humidity in the atmospheric case) for a particular location in space and time provide inputs to the parametrisation schemes. The tendencies in these variables over one time-step are computed by each scheme. The schemes are deterministic, local in horizontal space and time, but generally non-local in the vertical, and thus can be thought of as acting within a sub-grid column \citep[for a review, see][]{christensen2022}.

Deterministic parametrisation schemes typically assume that the grid box is large enough to contain many examples of the unresolved process, while simultaneously being `small enough to cover only a fraction of a large-scale disturbance' \citep{arakawa1974}. The parametrisation scheme is then tasked with computing the mean impact of a large ensemble of small-scale phenomena, all experiencing the same background state, onto the resolved scales: a well-posed problem. However, the Navier-Stokes equations are scale invariant \citep{lovejoy2013}, leading to the emergence of power-law behaviour in fluid flows, including in the atmosphere \citep{nastrom1985} and oceans \citep{storer2022}. In other words, fluid motions are observed at a continuum of scales and the spectral gap between resolved and unresolved scales required by deterministic parametrisation schemes does not exist. There will always be variability in the real flow on scales similar to the truncation scale, such that it is not possible to determine the impact of the sub-grid processes back on the grid-scale with certainty: the grid-scale variables cannot fully constrain the sub-grid motions. Deterministic parametrisations will always be a source of error in predictions. 

\begin{figure}[t]%
\includegraphics[width=0.95\textwidth]{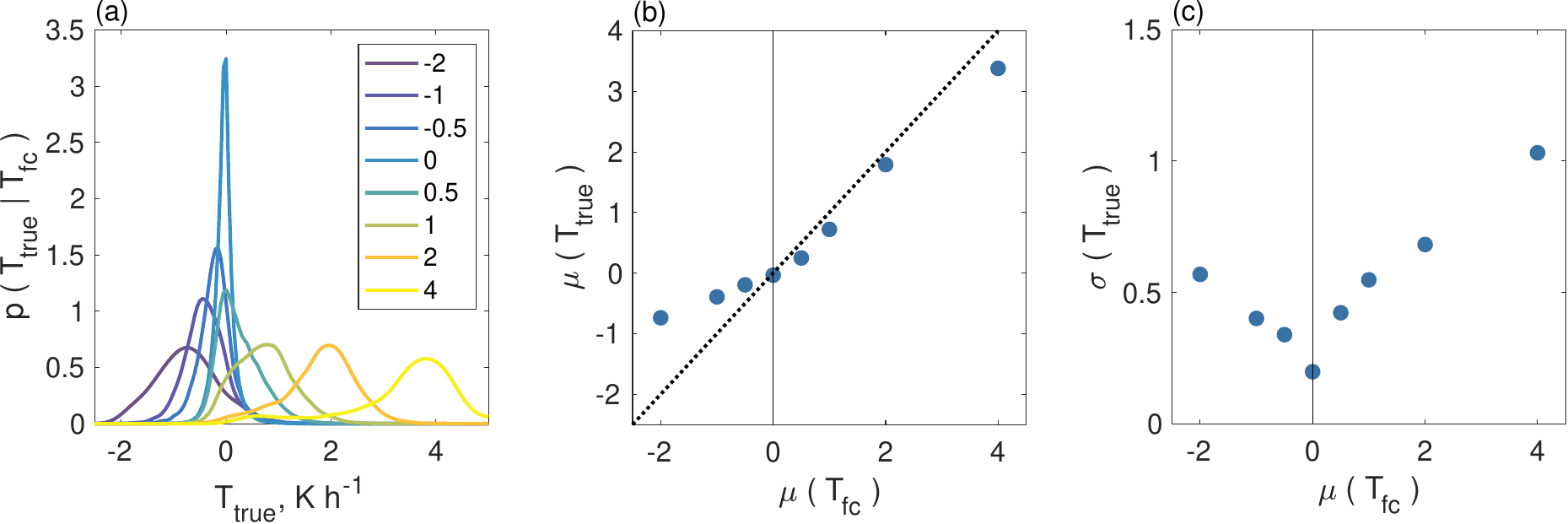}
{\caption{Coarse-graining studies provide evidence for stochastic parametrisations. (a), the pdf of `true' sub-grid temperature tendencies derived from a high-resolution simulation is conditioned on the tendency predicted by a deterministic forecast model ($T_{fc}$: colours in legend). (b) Mean `true' tendency conditioned on $T_{fc}$. For this forecast model, positive temperature tendencies are well calibrated, while negative temperature tendencies are biased cold. (c) Standard deviation of `true' tendency conditioned on $T_{fc}$. For this forecast model, the uncertainty in the `true' tendency increases with the magnitude of the low-resolution forecast tendency. Figure adapted from \citet{christensen2020}.}
\label{fig:CG}}
\end{figure}

Evidence supporting these theoretical considerations can be provided by coarse-graining studies. For example, \citet{christensen2020} takes a km-scale simulation that accurately captures small-scale variability as a reference. This simulation is coarse-grained to the resolution of a typical NWP model and used to initialise a low-resolution forecast model which uses deterministic parametrisation schemes. The true evolution of the coarse-grained high-resolution model is compared to that predicted by the forecast model. For a given deterministic forecast tendency, the high-resolution simulation shows a \emph{distribution} of tendencies, as shown in Figure~\ref{fig:CG}. While the mean of this distribution is captured by the parametrised model, there is substantial variability about the mean which is not captured.

Stochastic parametrisations provide an alternative paradigm to deterministic schemes. Instead of representing the mean sub-grid tendency, a stochastic approach represents one possible realisation of the sub-grid scale process \citep{arnold2013}. This means a stochastic scheme can be constructed to capture the variability observed in high-resolution datasets but missing from deterministic schemes \citep{buizza1999a}, as shown in Figure~\ref{fig:CG} . 

To design a stochastic parametrisation, a probability distribution must be specified which represents the distribution of possible sub-grid scale processes, which can be conditional on the observed scale state. A random number is then generated and used to draw from the modelled distribution. Importantly, spatio-temporal correlations can be included into the draw. The resultant parametrisation is no longer a sub-grid scheme, but a \emph{sur-grid} scheme, being non-local in space and time. This is better able to capture the scale-invariance of the Navier-Stokes equations \citep{Palmer_2019}. Evidence for the optimal spatio-temporal correlation scales can be provided by coarse graining studies \citep{christensen2020}.

We stress that it is impossible to know what the true sub-grid tendency would have been. A stochastic scheme acknowledges this. Different calls to the stochastic scheme will produce different tendencies, for example at different points in time in the model integration or in different realisations of an ensemble forecast. It is then possible to assess how the uncertainty in the parametrised sub-grid tendency interacts with the rest of the model physics, and ultimately leads to differences in the forecast. For this reason, we commonly refer to stochastic parametrisations as representing \emph{model uncertainty}.

\subsection{Benefits of stochastic parametrisations for weather and climate prediction}

In section~\ref{sec:closure}, we argued that deterministic parametrisation inevitably leads to errors in the predicted sub-grid tendencies, and that stochastic approaches allow us to capture the true variability in the impact of the sub-grid scales. It is important to highlight that this is not simply a theoretical nicety, but instead has large implications for the skill of the forecast model. This is because these small-scale errors in the forecast will not remain confined to the smallest scales. Instead, the chaotic nature of the atmosphere means errors will exponentially grow in time and cascade upscale in space \citep{lorenz1969}, causing model simulations to diverge substantially from the true atmosphere. 

Since errors in our forecasts are inevitable, instead of a single `best guess' prediction, operational centres typically make a set (or `ensemble') of equally likely weather forecasts \citep{bauer2015}. The goal is to produce a \emph{reliable} forecast, in which the observed evolution of the atmosphere is indistinguishable from individual ensemble members \citep{wilks2006}. To generate the ensemble, the initial conditions of the members are perturbed to represent uncertainties in estimates of the current state of the atmosphere. However, if only initial condition uncertainty is accounted for, the resultant ensemble is overconfident, such that the observations routinely fall outside of the ensemble forecast. A reliable forecast must also account for model uncertainty \citep{buizza1999a}. Stochastic parametrisations have transformed the reliability of initialised forecasts \citep{buizza1999a,weisheimer2014a,berner2017,ollinaho2017}, and so are widely used across the weather and subseasonal-to-seasonal forecasting communities. New developments in stochastic parametrisations can further improve the reliability of these forecasts \citep{Christensen2017b}

Climate prediction is a fundamentally different problem to weather forecasting \citep{lorenz1975}. It seeks to predict the response of the Earth-system to an external forcing (greenhouse-gas emissions). The goal is to predict the change in the statistics of the weather over the coming decades, as opposed to a specific trajectory. It has been shown that stochastic parametrisations from the weather forecasting community can substantially improve climate models \citep{berner2017,Christensen2017a,Strommen2019}. Including stochasticity in models can improve long-standing biases in the mean state, such as the distribution of precipitation \citep{Strommen2019}, as well as biases in climate variability, such as the El Niño–Southern Oscillation \citep{Christensen2017a,Yang2019}. Some evidence indicates that stochastic parametrisations can mimic the impacts of increasing the resolution of models \citep{dawson2015,Vidale2021}, likely by improving the representation of small-scale variability.

\subsection{Current stochastic approaches}

Stochastic parametrisations often couple with existing deterministic parametrisations. They can therefore be thought of as representing random errors in the deterministic scheme. One approach, the `Stochastically Perturbed Parametrisation tendencies' (SPPT) scheme, multiplies the sum of the output of the deterministic parametrisations by a spatio-temporally correlated random number with a mean of one \citep{buizza1999a,Sanchez2016,leutbecher2017}. An alternative approach takes parameters from within the parametrisations schemes and varies these stochastically to represent uncertainty in their values \citep[Stochastically Perturbed Parameters (SPP):][]{christensen2015d,ollinaho2017}. Both these approaches are holistic (treating all the parametrised processes), but make key assumptions about the nature of model error. Other approaches are more physically motivated. For example, the Plant-Craig scheme \citep{Plant2008} represents convective mass flux as following a Poisson distribution, motivated using ideas from statistical mechanics \citep{Craig2006}. The scheme predicts the convective mass flux at cloud base, which can be used as the closure assumption in an existing deterministic deep convection scheme. These ideas were subsequently extended for shallow convection by \citet{Sakradzija2016}.

A common conclusion across stochastic schemes is the need to include spatio-temporal correlations into the stochasticity in order for the scheme to have a significant impact on model skill. The need for correlations from a practical point of view complements the physical justification put forward in Section ~\ref{sec:closure}. Correlations are typically implemented using a first-order auto-regressive process in time, and using a spectral pattern in space \cite{christensen2015,johnson2019,palmer2009} . The correlation scale parameters can be tuned to maximise forecast skill.

\section{Machine Learning for Stochastic Parametrisation}\label{sec:ml}

\subsection{Why could ML be useful?}

Machine Learning (ML) solutions are a natural fit for stochastic parametrisation. There is a long history of data-driven approaches proposed by the stochastic parametrisation research community. For example, \citet{christensen2015d} use statistics derived from a data-assimilation approach to constrain the joint distribution of four uncertain parameters in a stochastic perturbed parameter scheme. Alternatively, the Markov Chain--Monte Carlo approaches of \citet{crommelin2008} and \citet{dorrestijn2013} begin by clustering high--fidelity data to produce a discrete number of realistic tendency profiles, before computing the conditional transition statistics between the states defined by the cluster centroids. On the other hand, \emph{operationally} used stochastic parametrisations (e.g. SPPT, SPP) are pragmatic, with only limited evidence for the structure that they assume. There is therefore substantial room for improvement. We note that the cost of existing stochastic parametrisations, such as SPPT, are generally low, so parametrisation improvement, not computational speed-up, is the principal goal here.


\subsection{Predicting the PDF}

Common to deterministic ML parametrisations, stochastic ML schemes must obey physical constraints, such as dependency on resolved-scale variables, correlations between sub-grid tendencies, and conservation properties. In addition, there are two further challenges unique to ML for stochastic parametrisation. The first is the need to learn the distribution which represents uncertainty in the process of interest. 

\begin{figure}[t]%
\includegraphics[width=0.55\textwidth]{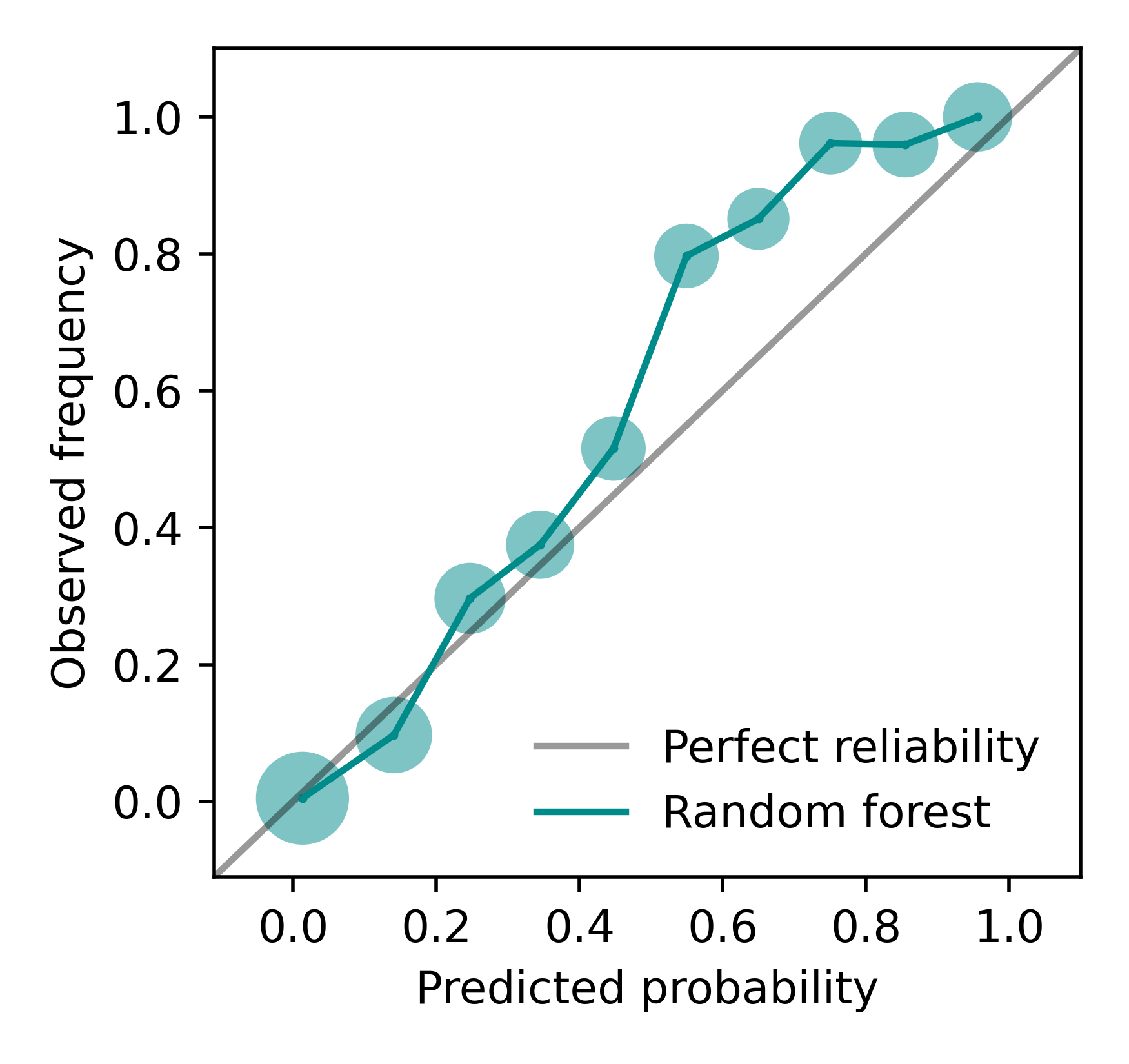}
{\caption{Reliability curve for convection occurrence estimated by the random forest (green line), which is close to perfect reliability (grey line). The random forest was developed for use as a stochastic convection trigger function. The circle sizes are proportional to the log of the number of samples per bin; there are many more non-convection events (91\%) than convection events (9\%). Figure adapted from \citet{miller2024}.}
\label{fig:trigger}}
\end{figure}

One approach is to couple a stochastic ML approach to an existing deterministic parametrisation. A sub-component of the existing scheme can be identified and the uncertainty in that component represented using an ML framework. For example, \citet{miller2024} develop a probabilistic ML approach to replace a deterministic convection trigger function (see also \citep{ukkonen2019}). A random forest was trained on large scale atmospheric variables, such as temperature and humidity, selected from the variational analysis over the Southern Great Plains (USA) observational site \citep{tang2019}. Since multiple sub-grid atmospheric states are possible for single large-scale state, the random forest is used to predict the probability of convection occurrence thereby capturing uncertainty in the convective trigger. Figure~\ref{fig:trigger} shows the reliability curve for the random forest estimates of convection occurrence. For this dataset, the random forest produced \emph{reliable} convection estimates (i.e. the predicted probabilities match the conditional observed frequency of convection) with minimal hyperparameter adjustment \citep{miller2024}. This is potentially because of the random forest’s feature bootstrapping and ensemble averaging methods. In addition, the random forest assigned a relatively even distribution of predicted probabilities greater than zero (Figure~\ref{fig:trigger}) indicating good \emph{resolution}. In general, both random forests and neural networks have been found to produce reliable probabilities without post-processing calibration, in comparison to other ML methods \citep{niculescu-mizil2005}. 

Instead of coupling to an existing parametrisation, an alternative approach is to replace a conventional parametrisation with an ML approach which generates samples from the sub-grid distribution, conditioned on the grid-scale variables. For example, \citet{guillaumin2021} model sub-grid momentum forcing using a parametric distribution, and train a neural network to learn the state-dependent parameters of this distribution \citep{guillaumin2021}. The distribution is then sampled from during model integration \citep{zhang2023implementation}. Alternatively, \citet{gagne2019machine} use a generative adversarial network (GAN) to generate samples conditioned on the resolved scale variables. When tested in online mode, the GAN was found to outperform baseline stochastic models and could produce reliable weather forecasts in a simple system. \citet{Nadiga2022} extend this work, and demonstrate that a GAN trained using atmospheric reanalysis data can generate realistic profiles in offline tests. However, it is not known whether a GAN truly learns the target distribution \citep{Arora_2018}. \citet{behrens2024improving} train a Variational Encoder-Decoder (VED) network to represent unresolved moist-physics tendencies derived from a superparametrised run with the Community Atmosphere Model: variability in the output profiles is generated by perturbing in the latent space of the encoder-decoder, giving improvements over a simple monte-carlo dropout approach. 

Learning to predict the sub-grid distribution is not sufficient for implementation of a stochastic ML scheme. The second half of the problem concerns how to draw from the predicted distribution. Implementations of ML stochastic parametrisations to date have not generally addressed this half of the problem. For example, \citet{zhang2023implementation} choose to use noise uncorrelated in space and time to implement the mesoscale eddy parametrisation of \citet{guillaumin2021} in the Modular Ocean Model. \citet{behrens2024improving} also use uncorrelated noise to sample from the VED network, which could explain the muted impact observed when using the scheme. In contrast, \citet{gagne2019machine} feed correlated noise into their GAN to draw from the generator, though the characteristics of this noise, including standard deviation and correlation statistics, were tuned according to forecast skill and not learnt from the data.

\subsection{Spatio-temporal correlations} \label{sec:corrs}

The second key challenge is developing an ML approach to capture the correlations in sub-grid tendencies across neighbouring columns in space and time, while still being practical to implement in a climate model. Any skilful ML approach must address this challenge.


Some early work has addressed temporal correlations. For example, \citet{Shin_2022} modelled entrainment into convective clouds as a stochastic differential equation with parameters predicted by NNs. The resultant solution is analogous to the first-order autoregressive models widely used in stochastic parametrisations, but for the case of continuous time. Alternatively, \citet{parthipan2023} use recurrent neural networks (RNN) in a probabilistic framework to learn the temporal characteristics of sub-grid tendencies in a toy atmospheric model. An RNN is a natural data-driven extension to simple autoregressive models: it can learn non-linear temporal associations, and also learn how many past states to use when predicting future tendencies. \citet{parthipan2023} found that using an RNN to model the temporal dependencies resulted in improved performance over a first order auto-regressive baseline \citep{gagne2019machine}. This indicates that more complex correlation structures may improve forecasts in atmospheric models. 

\begin{figure}[t]%
\includegraphics[width=0.95\textwidth]{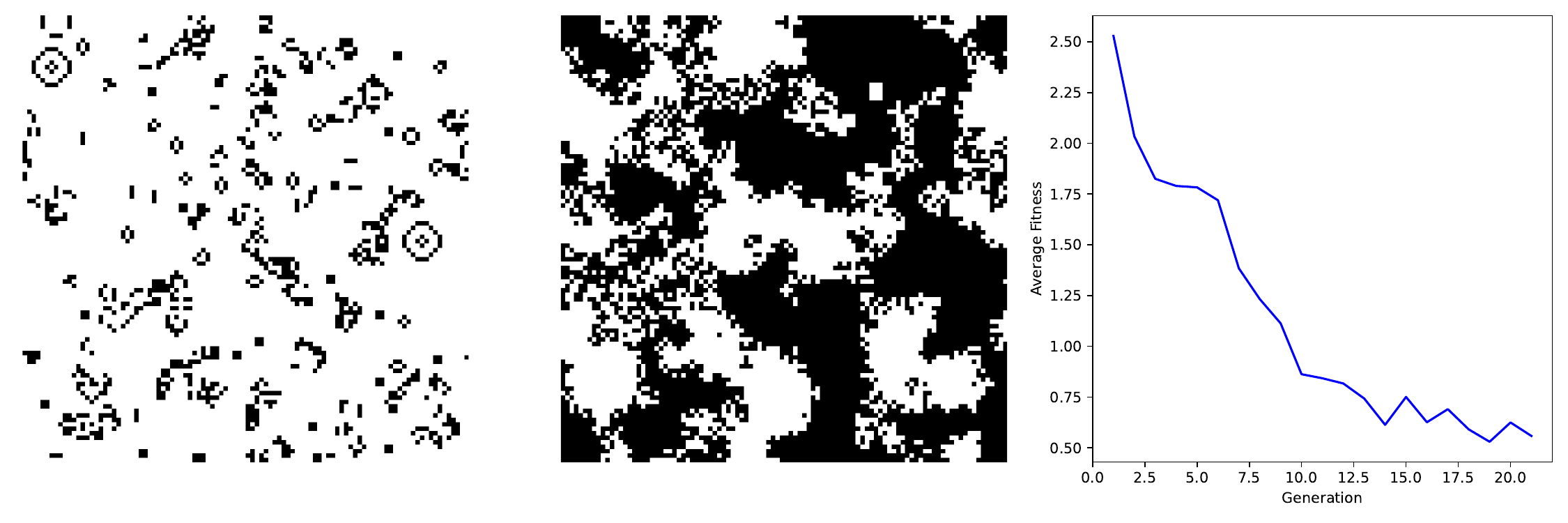}
{\caption{a. The classic cellular automata, the game of life, after 70 rule iterations on random initial conditions. b. A set of rules discovered through the use of a genetic algorithm after 70 iterations from a random initial condition. c. An example of fitness convergence for a genetic algorithm scheme.}
\label{fig:ca_ga}}
\end{figure}

Spatial correlations are arguably more difficult to address. This is because of the need to implement any scheme in a model where sub-grid columns are typically treated independently at integration stage, and the model is parallelised accordingly. However, inspiration can be taken from the approaches used in existing stochastic parametrisation schemes. For example, \citet{bengtsson2011large,bengtsson2013stochastic} have used a `game of life' cellular automaton (CA: Figure~\ref{fig:ca_ga}.a) to generate correlated fields for use in convection parametrisation. The CA operates on a grid finer than that used by the model parametrisations, such that each CA grid cell represents one convective element. The CA self-organises in space and persists in time, introducing spatio-temporal correlations which can be coupled to an existing deterministic convection scheme \citep{bengtsson2013stochastic}. However, the spatial correlations generated through this approach are extremely difficult to tune by hand (L. Bengtsson, pers. comm., 2018). One possible ML approach is to use a genetic algorithm (GA), which breeds and mutates successful rule sets, to navigate the large rule space. By using a fitness function based on the fractal dimension of the state after a certain number of model steps \citep[e.g. following][]{christensen2021_fractal}, a GA can discover cellular automata that produce a distribution of cells that more closely match those observed in clouds, as shown in Figure~\ref{fig:ca_ga}. Further work is needed to fully explore this possibility. 

An alternate approach to address spatial correlations incorporates information from neighbouring grid cells, as is utilised in the stochastic convection scheme of \citet{Plant2008} or in the deterministic Convolutional Neural Network proposed by \citet{wang2022}. Here accuracy gain from non-local inputs may be found to outweigh the increase in computational cost.


\subsection{Structure}

The question of finding the optimal structure for an ML model is common to all domains. Whilst tremendous success has come in the language domain from Transformer-based models \cite{vaswani2017attention}, the same tools have not similarly transformed the modelling of continuous processes (those that take real values). In fact, in the physical domain, there is an ongoing search for the most appropriate architecture. For example, the data-driven numerical weather prediction model GraphCast \cite{lam2022graphcast} uses a graph neural network as the backbone, whilst FourCastNet \cite{pathak2022fourcastnet}  and Pangu-Weather \cite{bi2023accurate} each develop separate inductive biases to use on top of a Transformer backbone. 

\begin{figure}[t]%
\includegraphics[width=0.95\textwidth]{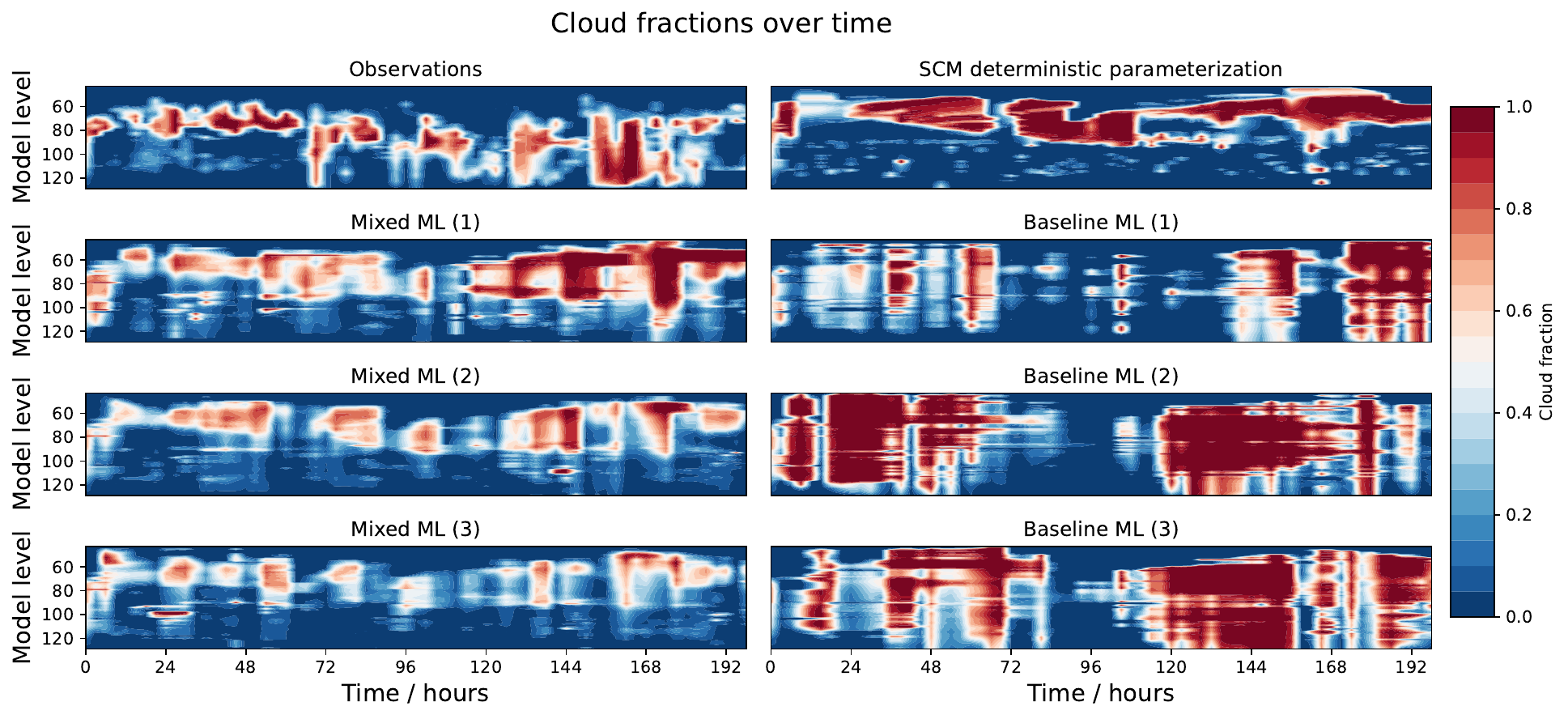}
{\caption{Cloud fractions as a function of height (model levels) for forecasts of 200 hours. Observed cloud fraction is compared to that from the operational deterministic parametrisation, and to two stochastic ML models. The Baseline ML model is a simple feed-forward neural network, whilst the Mixed ML model separates the task of modelling into a binary categorisation and continuous prediction problem. These are probabilistic models, and three sampled trajectories are shown for both. The mixed model is better able to create and remove cloud. Adapted from \citet{parthipan_thesis}.}
\label{fig:cloud_cover}}
\end{figure}

This problem is also encountered when building stochastic ML parametrisations. For example, developing a custom ML architecture was key when developing a probabilistic ML parametrisation for cloud cover for the European Centre for Medium-range Weather Forecasts single column model (SCM) \citep{parthipan_thesis}. It was found that a simple feed-forward neural network struggled to create and remove clouds appropriately. This is because transitions between cloud-containing and cloudless states are relatively rare in the training data. One solution to this problem was to use a mixed continuous-discrete model, which separated the task into i) predicting the probability of a cloud-containing state and ii) given a cloud-containing state, predicting the amount of cloud. Such models are often used for situations when the probability distribution has a discrete spike at the origin combined with a continuous positive distribution \cite{weld2017modeling}. Figure \ref{fig:cloud_cover} shows how this improvement in ML model structure results in substantial improvements in predicted cloud cover. However, this was a bespoke solution for the chosen problem. It is not clear whether a general architectural leap could bring enormous benefits to the field of physical modelling, as the Transformer did for language.

\section{Conclusions} \label{sec:conc}

Sub-grid parametrisations are a large source of error in weather and climate forecasts. Machine Learning (ML) is transforming this field, leading to reduced errors and large speed-ups in model predictions. However, a key widely-used assumption in ML parametrisation is that of determinism. We have argued that this assumption is flawed, and that substantial progress could be made by moving to a probabilistic framework. We discuss first attempts to develop such stochastic ML parametrisations and highlight remaining challenges including learning sur-grid correlation structures and suitable architectures. Fortunately, ample training data exists to learn such probabilistic parametrisations. The high-resolution datasets which are coarse-grained and used to train deterministic ML parametrisations are suitable for this task \citep[e.g.][]{brenowitz2020,yuval2020,Beucler2021climateinvariant}. In addition, a new multi-model training dataset is in production as part of the Model Uncertainty Model Intercomparison Project (MUMIP: \url{https://mumip.web.ox.ac.uk}), which will be ideal for ML stochastic parametrisations. Exploring this dataset will be the focus of future work.


\paragraph{Funding Statement}
HC was supported by UK Natural Environment Research Council grant number NE/P018238/1, and by the Leverhulme Trust Research Project Grant `Exposing the nature of model error in weather and climate models'.
SK and GM acknowledge support from the UK National Environmental Research Council Award NE/S007474/1. GM further acknowledges the University of Oxford Clarendon Fund. RP was funded by the Engineering and Physical Sciences Research Council grant number EP/S022961/1.

\paragraph{Competing Interests}
None

\paragraph{Data Availability Statement}
The coarse-grained data used in Figure~\ref{fig:CG} are archived at the Centre for Environmental Data Analysis (\url{http://catalogue.ceda.ac.uk/uuid/bf4fb57ac7f9461db27dab77c8c97cf2}). The ARM variational analysis dataset used to produce Figure~\ref{fig:trigger} can be downloaded from \url{https://www.arm.gov/capabilities/science-data-products/vaps/varanal}. The cloud fraction experiments in Figure~\ref{fig:cloud_cover} were run using the ECMWF OpenIFS single column model version 43r3, available for download from \url{https://confluence.ecmwf.int/display/OIFS}. 

\paragraph{Ethical Standards}
The research meets all ethical guidelines, including adherence to the legal requirements of the study country.

\paragraph{Author Contributions}
Conceptualisation, Supervision: HC, Investigation, Methodology, Software, Visualisation, Writing --- original draft, Writing --- review and editing: HC, SK, GM, RP. All authors approved the final submitted draft.

\bibliographystyle{plainnat}
\bibliography{Christensen}

\end{document}